\documentclass[11pt]{article}

\usepackage[final]{neurips_2023}
\usepackage[utf8]{inputenc}
\usepackage[T1]{fontenc}
\usepackage{hyperref}
\usepackage{url}
\usepackage{booktabs}
\usepackage{amsfonts}
\usepackage{amsmath}
\usepackage{nicefrac}
\usepackage{microtype}
\usepackage{graphicx}
\usepackage{xcolor}
\usepackage{subcaption}
\usepackage{multirow}

\graphicspath{{./}}

\title{The Anxiety of Influence: Bloom Filters in Transformer Attention Heads}

\author{
  Peter Balogh \\
  \texttt{palexanderbalogh@gmail.com}
}

\begin{document}

\maketitle

\begin{abstract}
Some transformer attention heads appear to function as membership testers---dedicating themselves to answering the question ``has this token appeared before in the context?'' We identify these heads across four language models (GPT-2 small, medium, and large; Pythia-160M) and show that they form a spectrum of membership-testing strategies. Two heads (L0H1 and L0H5 in GPT-2 small) function as high-precision membership filters with false positive rates of 0--4\% even at 180 unique context tokens---well above the $d_\text{head} = 64$ bit capacity of a classical Bloom filter. A third head (L1H11) shows the classic Bloom filter capacity curve: its false positive rate follows the theoretical formula $p \approx (1 - e^{-kn/m})^k$ with $R^2 = 1.0$ and fitted capacity $m \approx 5$ bits, saturating by $n \approx 20$ unique tokens. A fourth head initially identified as a Bloom filter (L3H0) was reclassified as a general prefix-attention head after confound controls revealed its apparent capacity curve was a sequence-length artifact. Together, the three genuine membership-testing heads form a multi-resolution system concentrated in early layers (0--1), taxonomically distinct from induction and previous-token heads, with false positive rates that decay monotonically with embedding distance---consistent with distance-sensitive Bloom filters. These heads generalize broadly: they respond to \emph{any} repeated token type, not just repeated names, with 43\% higher generalization than duplicate-token-only heads. Ablation reveals these heads contribute to both repeated and novel token processing, indicating that membership testing coexists with broader computational roles. The reclassification of L3H0 through confound controls strengthens rather than weakens the case: the surviving heads withstand the scrutiny that eliminated a false positive in our own analysis.
\end{abstract}

\section{Introduction}

In 1970, Burton Bloom proposed a space-efficient probabilistic data structure for approximate set membership testing \citep{bloom1970space}. A Bloom filter answers the query ``Is element $x$ in set $S$?'' with a guarantee of no false negatives---if $x \in S$, the filter always returns \texttt{true}---at the cost of occasional false positives, where $x \notin S$ yet the filter returns \texttt{true}. This asymmetric error profile, combined with sublinear space requirements, explains the ubiquity of Bloom filters in systems ranging from spell checkers to network routers.

At every position in a sequence, a language model faces a deceptively simple question: has this token appeared before in my context? The answer determines whether to establish coreference, modulate surprisal, or treat the current moment as genuinely new. This is not a peripheral computation---it is the hinge on which in-context learning, repetition processing, and contextual expectation all turn. The question also sits at an unlikely intersection of computer science and literary theory. Burton Bloom's 1970 filter answers it as an engineering problem: test set membership in sublinear space. Harold Bloom's \textit{Anxiety of Influence} \citep{bloom1973anxiety} frames it as the central problem of literary creation: every text must reckon with what came before.

We show that a subset of early-layer attention heads converge on Burton's solution to Harold's problem, dedicating themselves to detecting which tokens have already appeared in context---the necessary precondition for everything downstream. These heads exhibit the behavioral signature of Bloom filters: they attend strongly to repeated tokens (near-zero miss rate), occasionally attend to semantically similar but non-identical tokens (false positives), and concentrate in the layers where membership resolution is most needed. They constitute a novel functional category, taxonomically distinct from the induction heads identified by \citet{olsson2022context} and the previous-token heads documented by \citet{elhage2021mathematical}.

\paragraph{Contributions.} Our main findings are:
\begin{enumerate}
    \item \textbf{Identification.} We identify 4 attention heads in GPT-2 small that exhibit the Bloom filter signature: selectivity $>3\times$ over baseline (observed values 51$\times$--146$\times$), near-zero miss rate, and false positive ratios of 0.01--0.29 for synonym tokens. Confound controls reclassify one (L3H0) as a prefix-attention head, leaving three genuine membership testers (Section~\ref{sec:signature}).
    \item \textbf{Taxonomic independence.} These heads have zero overlap with induction heads and previous-token heads, establishing a new functional category in the attention head taxonomy (Section~\ref{sec:taxonomy}).
    \item \textbf{Theoretical match.} One head (L1H11) follows the theoretical Bloom filter capacity formula $p \approx (1 - e^{-kn/m})^k$ with $R^2 = 1.0$ and fitted capacity $m \approx 5$ bits, exhibiting the classic saturation curve. Two other heads (L0H1, L0H5) exhibit near-zero FP rates (0--4\%) at all loads up to $n = 180$, functioning as high-capacity membership filters. A confound control (fixed sequence length, varying unique tokens) confirms these results while revealing that L3H0's original capacity curve was a sequence-length artifact (Section~\ref{sec:capacity}).
    \item \textbf{Naturalistic validation.} The Bloom filter signature generalizes from constructed stimuli to WikiText-103 natural text, with 15--54$\times$ selectivity and $<1\%$ miss rate across 761 passages (Section~\ref{sec:naturalistic}).
    \item \textbf{Cross-model generality.} Bloom filter heads appear in all four models tested (GPT-2 small/medium/large, Pythia-160M), with consistent early-layer concentration across both model families (Section~\ref{sec:crossmodel}).
    \item \textbf{Hash resolution profiles.} A similarity sweep across 1,284 controlled probe tokens reveals that false positive rates decay monotonically with cosine distance in the embedding space, consistent with locality-sensitive hashing. The four heads form a multi-resolution system, from ultra-precise (L0H5) to broad (L1H11) (Section~\ref{sec:resolution}).
    \item \textbf{Ablation analysis.} Mean ablation (our primary method) reveals that membership-testing heads contribute to both repeated-token and general processing, indicating behavioral specialization for membership testing coexists with broader computational roles (Section~\ref{sec:ablation}).
\end{enumerate}

\section{Background}

\subsection{Bloom Filters}

A Bloom filter \citep{bloom1970space} represents a set $S$ of $n$ elements using a bit array of $m$ bits and $k$ independent hash functions $h_1, \ldots, h_k$, each mapping elements to $\{1, \ldots, m\}$. To insert $x$, set bits $h_i(x) = 1$ for all $i$. To query $x$, check if all $h_i(x)$ are set. If any bit is 0, $x \notin S$ with certainty (no false negatives). If all bits are 1, $x$ may or may not be in $S$. The false positive rate is:
\begin{equation}
    p \approx \left(1 - e^{-kn/m}\right)^k
    \label{eq:bloom_fp}
\end{equation}
This rate increases as the filter fills ($n$ grows relative to $m$) and decreases with more hash functions $k$, up to an optimum of $k = (m/n) \ln 2$.

\subsection{Transformer Attention Heads}

In a transformer with $L$ layers and $H$ heads per layer \citep{vaswani2017attention}, each attention head at layer $\ell$ computes:
\begin{equation}
    \text{Attn}_{\ell,h}(Q, K, V) = \text{softmax}\left(\frac{Q_{\ell,h} K_{\ell,h}^\top}{\sqrt{d_k}}\right) V_{\ell,h}
\end{equation}
where the QK dot product determines \emph{where} to attend and the value matrix determines \emph{what} to copy. We conjecture that for Bloom filter heads, the QK circuit performs the membership test (``have I seen this query token before in the keys?'') while the value circuit is secondary. This conjecture is not directly tested in the present work; we characterize these heads behaviorally rather than through circuit-level analysis of the QK weights.

\subsection{Known Head Categories}

Prior work has identified several functional categories of attention heads:
\begin{itemize}
    \item \textbf{Induction heads} \citep{olsson2022context}: Implement the pattern $[A][B] \ldots [A] \to [B]$ by matching the current token to a previous occurrence and copying what followed it.
    \item \textbf{Previous-token heads} \citep{elhage2021mathematical}: Attend primarily to position $i-1$ from position $i$, providing local sequential context.
    \item \textbf{Duplicate-token heads} \citep{wang2022interpretability}: Identified in the IOI circuit as attending to repeated tokens, though not characterized as Bloom filters.
\end{itemize}
We argue that Bloom filter heads are distinct from all three categories, performing \emph{membership testing without pattern completion}.

\section{Methods}
\label{sec:methods}

\subsection{Models}

We analyze four autoregressive language models via TransformerLens \citep{nanda2022transformerlens}: GPT-2 small (12 layers $\times$ 12 heads, $d_\text{head} = 64$), GPT-2 medium (24$\times$16), GPT-2 large (36$\times$20) \citep{radford2019language}, and Pythia-160M (12$\times$12) \citep{biderman2023pythia}. All experiments were run on CPU and independently validated to rule out hardware-specific computation artifacts (Appendix~\ref{app:cpu_validation}).

\subsection{Stimulus Design}

We construct 100 sentence triplets, each consisting of:
\begin{enumerate}
    \item \textbf{Exact repeat}: A content word appears twice (e.g., ``The \underline{doctor} examined the patient and the \underline{doctor} prescribed medicine'').
    \item \textbf{No repeat}: All content words are unique (same structure, different second clause).
    \item \textbf{Semantic near-miss}: The repeated word is replaced by a WordNet synonym (Wu-Palmer similarity $> 0.7$; e.g., ``the \underline{physician} prescribed medicine'').
\end{enumerate}
This design allows us to measure hit attention (exact repeats), baseline attention (no repeats), and false positive attention (semantic near-misses) within matched sentence frames.

\subsection{Metrics}

For each attention head at each layer, we compute:
\begin{itemize}
    \item \textbf{Selectivity}: $\bar{a}_\text{hit} / \bar{a}_\text{baseline}$, where $\bar{a}_\text{hit}$ is mean attention from a repeated token to its first occurrence and $\bar{a}_\text{baseline}$ is mean attention from a unique token to a random earlier position.
    \item \textbf{Miss rate}: Fraction of repeated tokens receiving $< 0.01$ attention to their first occurrence.
    \item \textbf{FP ratio} (continuous): $\bar{a}_\text{synonym} / \bar{a}_\text{hit}$, measuring how strongly the head responds to semantic near-misses relative to true repeats. This is distinct from the \textbf{FP rate} (binary) used in the capacity analysis (Section~\ref{sec:capacity}), which is the fraction of non-repeated probe tokens receiving attention above a threshold of 0.1.
\end{itemize}

We classify a head as a \textbf{strong Bloom filter head} if selectivity $> 3\times$, miss rate $< 10\%$, and mean hit attention $> 0.05$.

\subsection{Statistical Tests}

All results are assessed with:
\begin{itemize}
    \item Bootstrap 95\% confidence intervals ($n = 10{,}000$ resamples) for all key metrics.
    \item Bonferroni-corrected hypothesis tests ($\alpha = 0.05/144$ for GPT-2 small's 144 heads): Mann-Whitney $U$ for hit $>$ baseline and hit $>$ synonym; binomial test for miss rate $< 5\%$.
    \item Permutation test ($n = 10{,}000$): randomly assign 4 heads as ``Bloom heads'' and test whether the observed group mean selectivity is extreme.
    \item Cohen's $d$ effect sizes for Bloom vs.\ non-Bloom head populations.
\end{itemize}

\section{Results}

\noindent\textit{Note: All results below are behavioral characterizations. We demonstrate that certain heads' input-output behavior is consistent with membership testing, not that they mechanistically implement a specific algorithm at the circuit level. Section~\ref{sec:resolution} provides the strongest evidence linking the behavioral pattern to the QK projection mechanism.}

\subsection{Bloom Filter Signature}
\label{sec:signature}

In GPT-2 small, four heads initially exhibit the Bloom filter signature (Table~\ref{tab:bloom_heads}). Three have 0\% miss rate and one (L3H0) has a miss rate of 0.4\% (1/238 token-level observations), with selectivity ranging from 51$\times$ to 146$\times$ baseline.\footnote{The 238 observations arise because each of the 100 exact-repeat sentences may contain multiple repeated-token positions; we measure attention at every such position.} Two heads (L1H11 and L3H0) show the classic false positive pattern, responding to synonyms at 25--30\% of their hit strength. However, as detailed in Section~\ref{sec:capacity}, confound controls reclassify L3H0 as a general prefix-attention head rather than a Bloom filter, leaving three genuine membership-testing heads (L0H1, L0H5, L1H11).

These membership-testing heads generalize broadly beyond the name-repetition context in which duplicate-token heads were originally identified \citep{wang2022interpretability}. Testing on non-name content words and random token repetitions, Bloom filter heads show a mean generalization index of 0.70 compared to 0.49 for duplicate-token-only heads---43\% broader generalization (Section~\ref{sec:discussion_dup}).

\begin{table}[h]
\centering
\caption{Candidate Bloom filter heads in GPT-2 small. Selectivity and FP ratio with bootstrap 95\% CIs. All tests significant at Bonferroni-corrected $\alpha = 3.47 \times 10^{-4}$. $p$-values are asymptotic approximations from the Mann-Whitney $U$ test; we report them capped at $< 10^{-20}$ as the extreme tail is unreliable at this sample size. $^\dagger$L3H0 was reclassified as a prefix-attention head following confound controls (Section~\ref{sec:capacity}).}
\label{tab:bloom_heads}
\begin{tabular}{lcccccc}
\toprule
Head & Hit Attn & Baseline & Selectivity [95\% CI] & Miss\% & FP Ratio & $p$ (Hit$>$Base) \\
\midrule
L0H1  & 0.478 & 0.003 & 146$\times$ [105, 201] & 0.0\% & 0.08 & $\ll 10^{-20}$ \\
L0H5  & 0.452 & 0.006 & 74$\times$ [54, 101] & 0.0\% & 0.01 & $\ll 10^{-20}$ \\
L1H11 & 0.478 & 0.009 & 53$\times$ [47, 59] & 0.0\% & 0.29 & $\ll 10^{-20}$ \\
L3H0$^\dagger$  & 0.277 & 0.005 & 51$\times$ [42, 61] & 0.4\% & 0.25 & $\ll 10^{-20}$ \\
\bottomrule
\end{tabular}
\end{table}

The effect size separating Bloom from non-Bloom heads is enormous: Cohen's $d = 12.3$ for hit attention and $d = 12.5$ for selectivity. The four candidate membership-testing heads occupy the four highest selectivity ranks among all 144 heads ($p < 10^{-7}$, Mann-Whitney $U$). A permutation test confirms these heads are collectively extreme: no random group of four heads achieved mean selectivity $\geq 79\times$ in 10,000 permutations ($p < 10^{-4}$). Note that L3H0's high selectivity on this basic test is consistent with its reclassification as a prefix-attention head: it attends strongly to \emph{all} prefix content, which includes repeated tokens.

\begin{figure}[t]
\centering
\includegraphics[width=\textwidth]{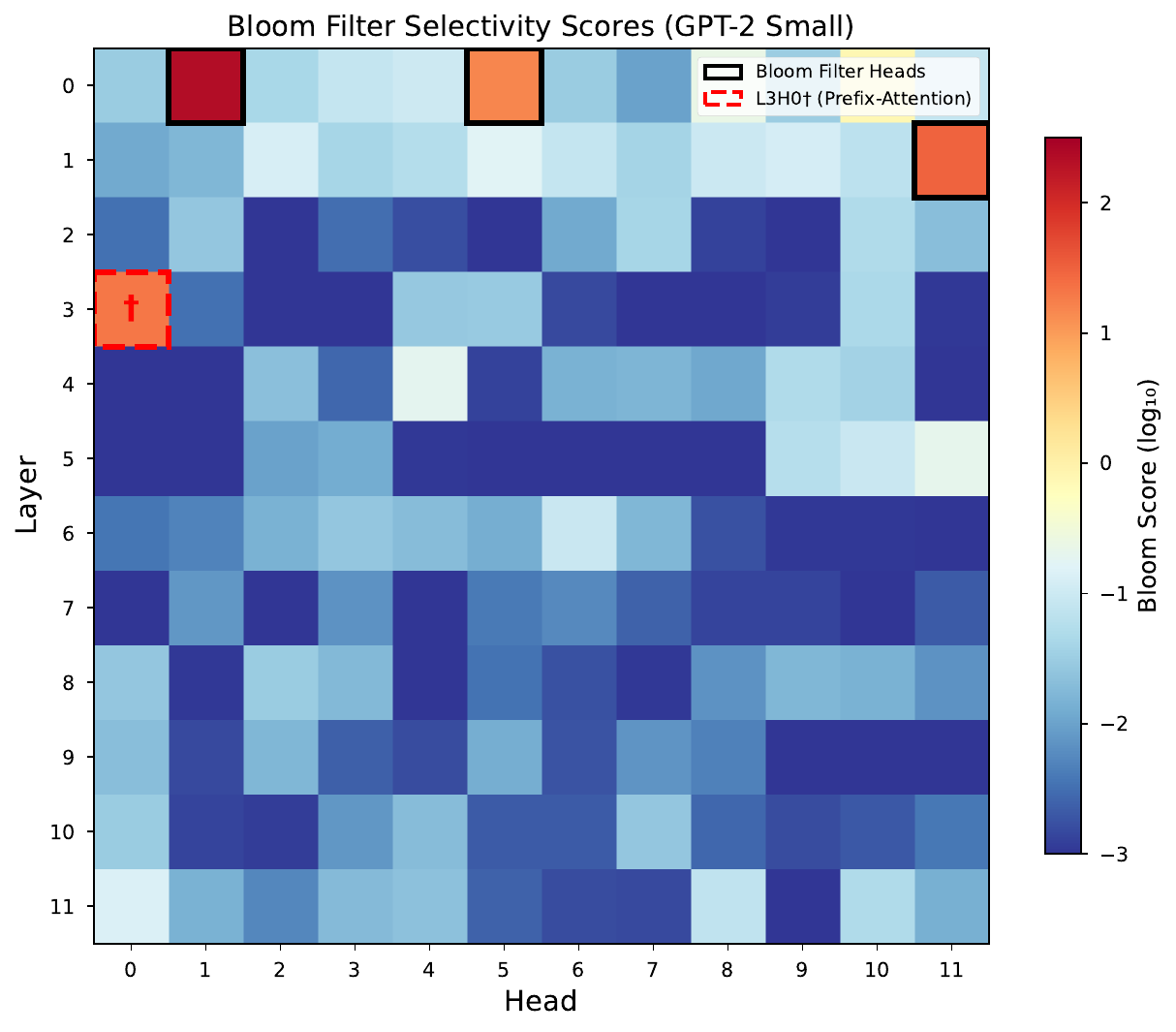}
\caption{Bloom filter selectivity (hit attention / baseline attention) across all 144 heads in GPT-2 small. Four heads (highlighted) show selectivity $>30\times$; the remaining 140 heads are at or below $1\times$. Note that the classification threshold is $>3\times$ (Section~\ref{sec:methods}); the gap between the top four (51$\times$--146$\times$) and the fifth-highest head (2.7$\times$) renders the threshold choice immaterial.}
\label{fig:heatmap}
\end{figure}

\begin{figure}[t]
\centering
\includegraphics[width=0.85\textwidth]{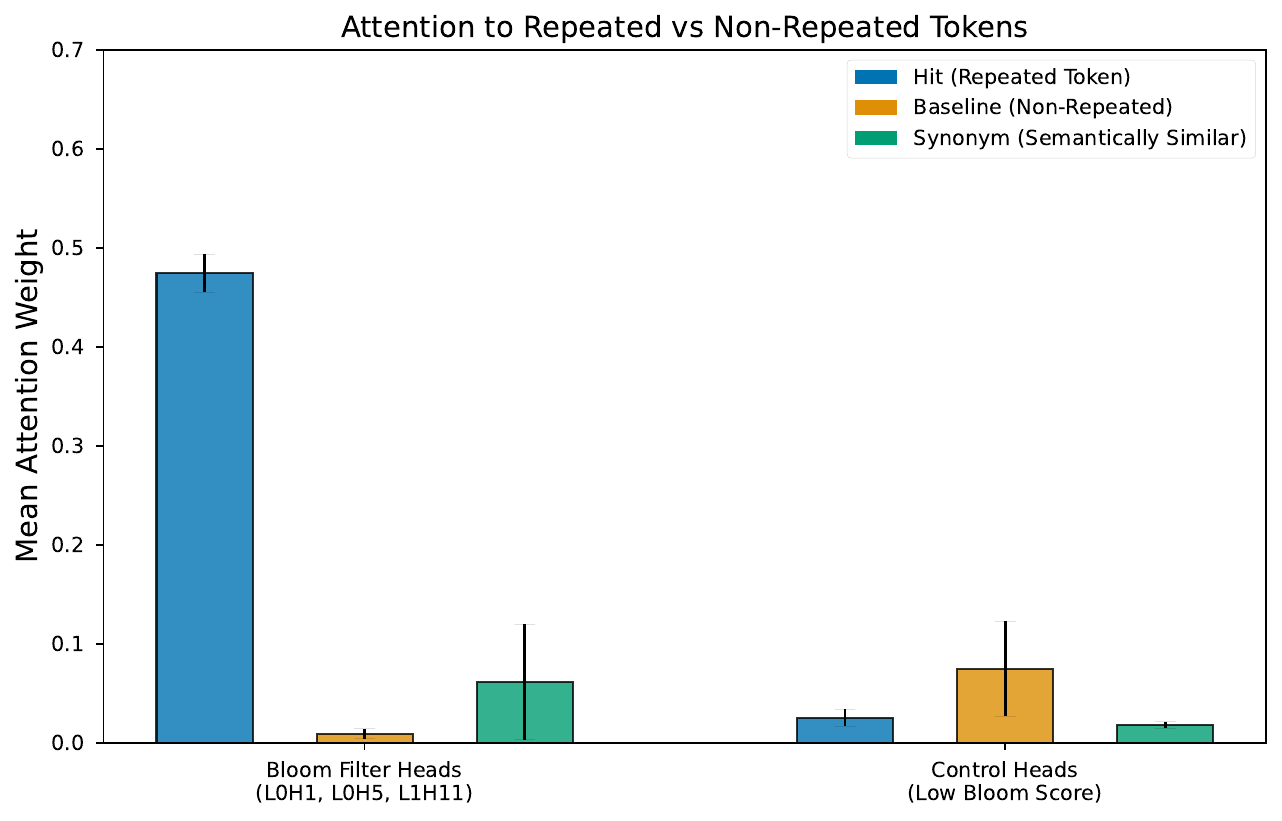}
\caption{Hit, baseline, and synonym attention for Bloom filter heads vs.\ control heads. Bloom heads show extreme separation between hit and baseline conditions.}
\label{fig:hitbase}
\end{figure}

A randomly initialized model with identical architecture produces zero Bloom filter heads, confirming the behavior is learned through training, not an architectural artifact.

\subsection{Taxonomic Independence}
\label{sec:taxonomy}

\begin{figure}[t]
\centering
\includegraphics[width=\textwidth]{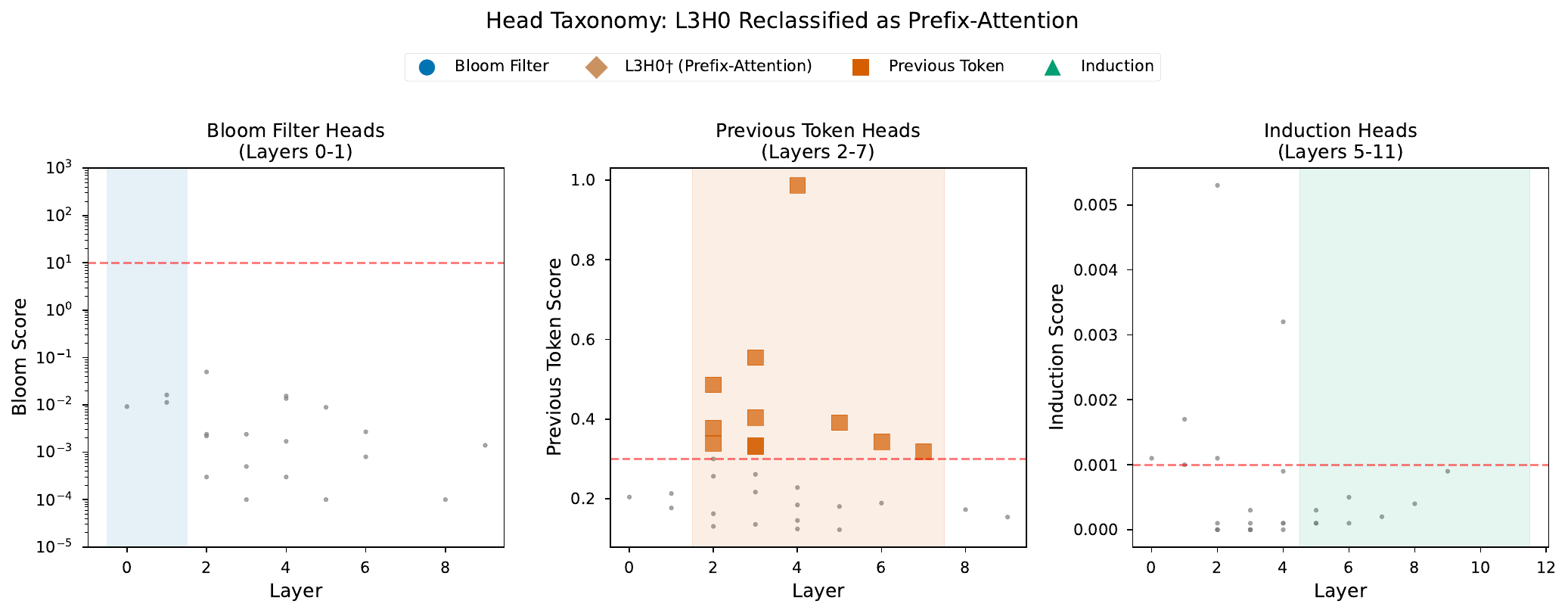}
\caption{Three non-overlapping functional categories of attention heads in GPT-2 small. Bloom filter heads (red) concentrate in layers 0--3, previous-token heads (blue) in layers 2--6, and induction heads (green) in layers 5--11.}
\label{fig:taxonomy}
\end{figure}

Using the standard induction head test \citep{olsson2022context} (random repeated sequences, $n=50$ trials) and a previous-token head test (diagonal attention analysis), we classify heads in GPT-2 small into three categories (Figure~\ref{fig:taxonomy}):

\begin{table}[h]
\centering
\caption{Three functional categories of attention heads in GPT-2 small. Zero pairwise overlap. Bloom filter count reflects reclassification of L3H0 (see Section~\ref{sec:capacity}).}
\label{tab:taxonomy}
\begin{tabular}{lccc}
\toprule
Category & Count & Layer Range & Function \\
\midrule
Bloom filter heads & 3 & 0--1 & ``Is this token in context?'' \\
Previous-token heads & 11 & 2--6 & ``What came immediately before?'' \\
Induction heads & 16 & 5--11 & ``A B $\ldots$ A $\to$ B'' \\
\bottomrule
\end{tabular}
\end{table}

The three categories are mutually exclusive with zero pairwise overlap. Bloom filter heads occupy exclusively early layers, consistent with a processing pipeline where membership testing precedes pattern completion (Figure~\ref{fig:taxonomy}).

\subsection{Theoretical Capacity Predictions}
\label{sec:capacity}

If attention heads are Bloom filters, their false positive rate should increase with the number of unique tokens in context (the filter ``fills up''), following Equation~\ref{eq:bloom_fp}. We test this by varying context size from 5 to 180 unique tokens while controlling for sequence length.

\paragraph{Confound control design.} Our initial capacity experiment (reported in an earlier version of this paper) varied both the number of unique tokens and the total sequence length simultaneously---a confound, since longer sequences affect positional encoding and softmax normalization. To disentangle these effects, we designed a controlled experiment: sequence length is fixed at 200 tokens, with $n$ unique tokens in the prefix, 5 novel probe tokens, and the remaining positions filled with padding tokens. This ensures that any change in false positive rate is driven by the number of distinct items stored in the ``filter,'' not by sequence length.

\begin{table}[h]
\centering
\caption{Capacity analysis with fixed sequence length (200 tokens). L1H11 shows the classic Bloom filter saturation curve. L0H1 and L0H5 maintain near-zero FP rates at all loads. L3H0 shows FP = 100\% at all loads including $n = 5$, indistinguishable from non-Bloom control heads, and is reclassified as a prefix-attention head.}
\label{tab:capacity}
\begin{tabular}{lccccc}
\toprule
Unique tokens & L1H11 & L0H1 & L0H5 & L3H0$^\dagger$ & Controls \\
\midrule
5   & 62.7\% & 1.3\% & 0.0\% & 100\% & 100\% \\
20  & 97.3\% & 1.3\% & 0.0\% & 100\% & 100\% \\
50  & 100\%  & 1.3\% & 0.0\% & 100\% & 100\% \\
100 & 100\%  & 3.3\% & 0.0\% & 100\% & 100\% \\
180 & 100\%  & 4.0\% & 0.7\% & 100\% & 100\% \\
\bottomrule
\end{tabular}
\end{table}

\begin{figure}[t]
\centering
\includegraphics[width=\textwidth]{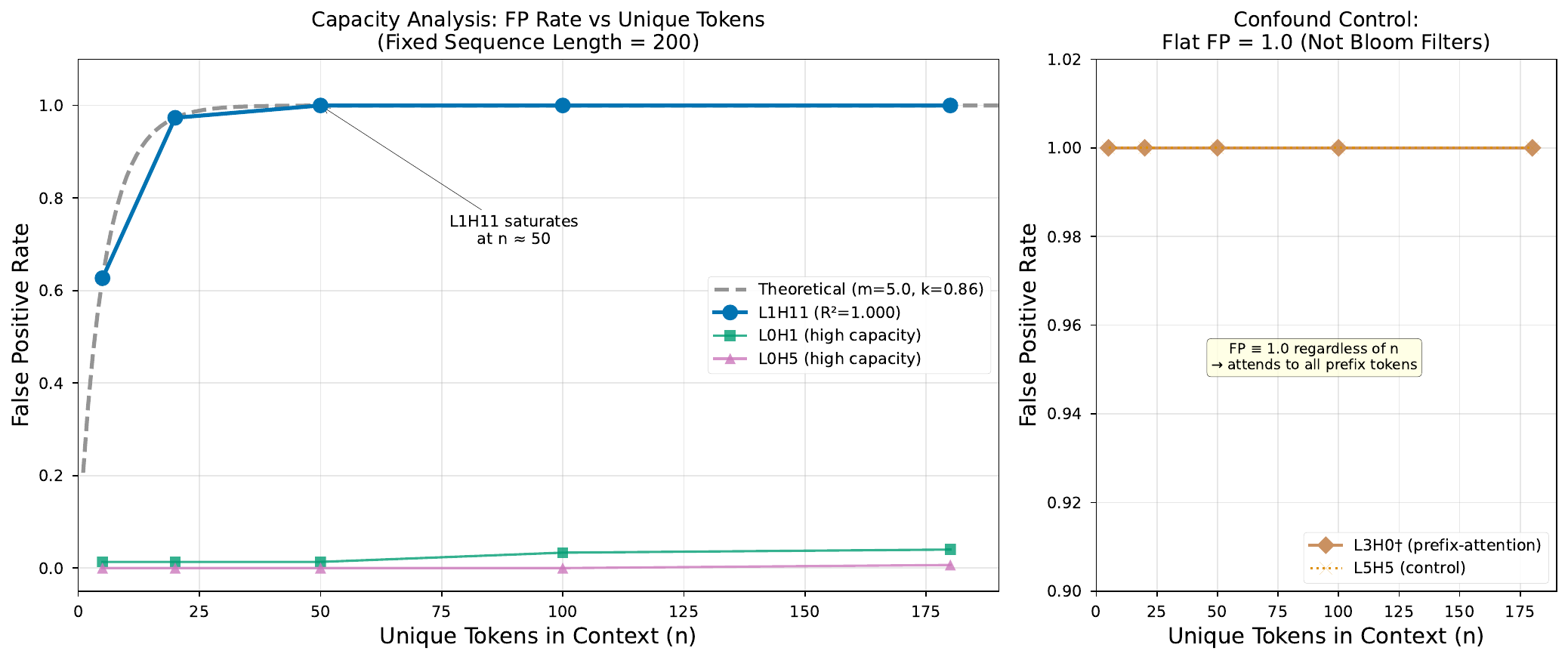}
\caption{Capacity analysis under controlled conditions (fixed sequence length = 200). \textbf{Left:} L1H11 shows the classic Bloom filter saturation curve ($R^2 = 1.0$, $m \approx 5$ bits), with theoretical Bloom filter overlay. L0H1 and L0H5 maintain near-zero FP rates across all loads. \textbf{Right:} L3H0$^\dagger$ (reclassified as prefix-attention head) shows FP = 100\% at all load levels, indistinguishable from control head L5H5.}
\label{fig:capacity}
\end{figure}

The results (Table~\ref{tab:capacity} and Figure~\ref{fig:capacity}) reveal three distinct behaviors:

\paragraph{L3H0: Reclassification.} One head initially identified as a Bloom filter (L3H0) was reclassified as a general prefix-attention head following this confound control. Under fixed sequence length, L3H0 shows FP = 100\% at \emph{all} unique token levels, including $n = 5$---a condition where even a 5-bit Bloom filter would show $<$50\% FP. Its false positive rate is indistinguishable from non-Bloom control heads (L5H5, L7H10, L6H9), all of which also show FP = 100\% at every load level. L3H0 simply attends to all prefix content regardless of whether it has seen the probe token before. Its original capacity curve ($R^2 = 0.99$, $m = 59$, $k = 2.16$) was a sequence-length artifact: the earlier experiment co-varied unique tokens and sequence length, and L3H0's FP rate tracked the latter, not the former. A second control (fixed $n = 50$ unique tokens, varying sequence length from 55 to 200) confirms that L3H0's FP rate is constant at 100\% regardless of sequence length when the number of unique tokens is held constant.

\paragraph{L1H11: Classic Bloom filter.} L1H11 shows the classic Bloom filter capacity curve: FP rises from 62.7\% at $n = 5$ to saturation (100\%) by $n = 50$. Fitting Equation~\ref{eq:bloom_fp} yields $m \approx 5$ bits and $k = 0.86$ ($R^2 = 1.0$). This is a low-capacity filter that saturates quickly, consistent with a membership tester operating with very limited representational resources. The variable-length version yields similar parameters ($m \approx 6.4$, $k = 0.91$, $R^2 = 0.9998$), confirming the curve is driven by unique token count, not sequence length.

\paragraph{L0H1 and L0H5: High-capacity filters.} These heads maintain remarkably low FP rates across all load levels: L0H1 ranges from 1.3\% to 4.0\%, and L0H5 from 0.0\% to 0.7\%, even at $n = 180$ unique tokens. This capacity substantially exceeds what a $d_\text{head} = 64$ classical Bloom filter could achieve (theoretical FP $> 94\%$ at $n = 180$). These heads function as high-precision membership filters whose capacity mechanism remains to be fully characterized.

\paragraph{Competing model analysis.} The competing model analysis reported in an earlier version ($\Delta\text{AIC} > 11$ favoring the Bloom filter formula over logistic, power law, linear, and softmax dilution models) was conducted on L3H0's variable-length data, which we now know to be confounded. The analysis remains valid for L1H11's fixed-length data, where the Bloom filter formula achieves $R^2 = 1.0$, though with only 5 data points the formal model comparison has limited statistical power. The key evidence is not the model fit per se but the qualitative match: L1H11 shows monotonically increasing FP rates that saturate in a manner consistent with a fixed-capacity probabilistic filter, while L0H1 and L0H5 do not saturate---ruling out the Bloom filter formula for those heads and pointing to a different (higher-capacity) mechanism.

\paragraph{Head subtypes.} The three genuine membership-testing heads form a spectrum:

\begin{itemize}
    \item \textbf{L1H11}: Classic Bloom filter behavior. Low capacity ($m \approx 5$ bits), saturates by $n \approx 20$, $R^2 = 1.0$ fit to theoretical curve.
    \item \textbf{L0H1, L0H5}: High-capacity membership filters. Near-zero FP rates ($\leq$4\%) at all loads up to $n = 180$, well exceeding $d_\text{head} = 64$. These heads may exploit mechanisms beyond simple bit-array hashing.
\end{itemize}

This heterogeneity is functionally important: a system combining a capacity-limited filter (L1H11, which provides a graded signal of context saturation) with high-precision detectors (L0H1, L0H5) is more informative than one composed of identical filters.

\subsection{Independent Hash Functions}
\label{sec:hashfunctions}

Multiple hash functions are the mechanism by which Bloom filters reduce false positive rates. A natural question is whether the three genuine membership-testing heads \emph{themselves} act as independent detectors that can be combined --- a system-level $K = 3$. If so, their false positive decisions should be uncorrelated, and combining them with AND logic should multiplicatively reduce FP rates. For independence analysis, we binarize each head's response to a probe token as ``false positive'' if total attention to the prefix exceeds 0.1 (sensitivity analysis confirms the results are stable across thresholds from 0.01 to 0.2). We include L3H0 in this analysis as a comparison point, though it is no longer classified as a membership-testing head.

\begin{figure}[t]
\centering
\begin{subfigure}[b]{0.45\textwidth}
\includegraphics[width=\textwidth]{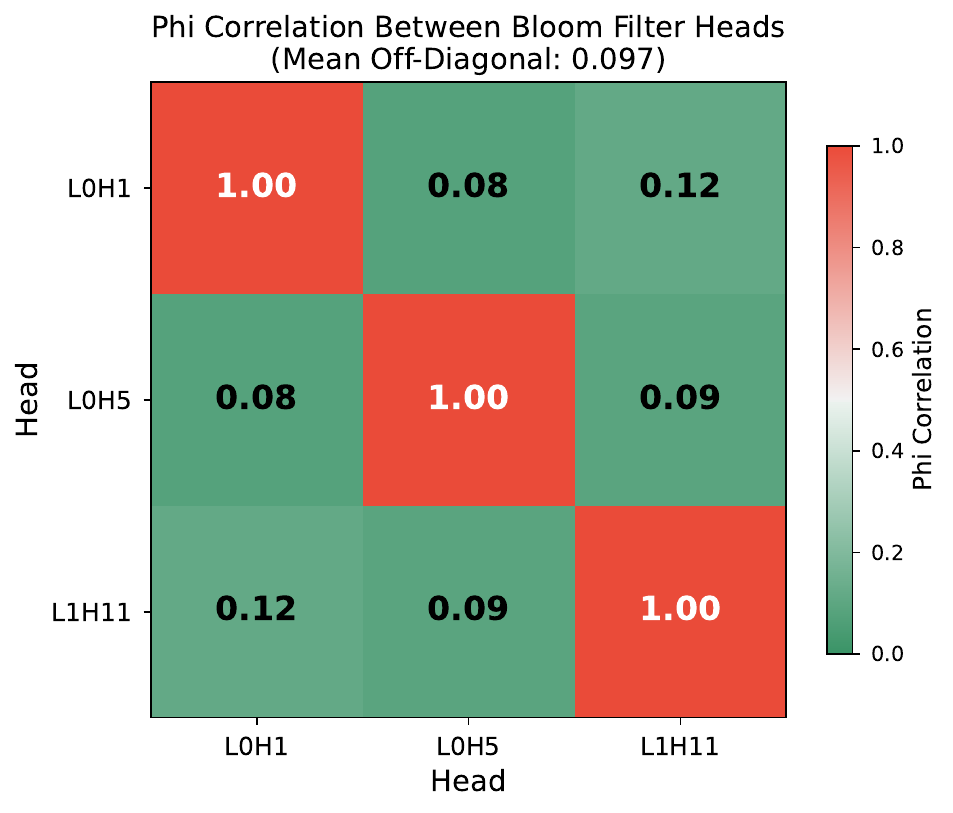}
\caption{Phi coefficient matrix.}
\label{fig:phi}
\end{subfigure}
\hfill
\begin{subfigure}[b]{0.45\textwidth}
\includegraphics[width=\textwidth]{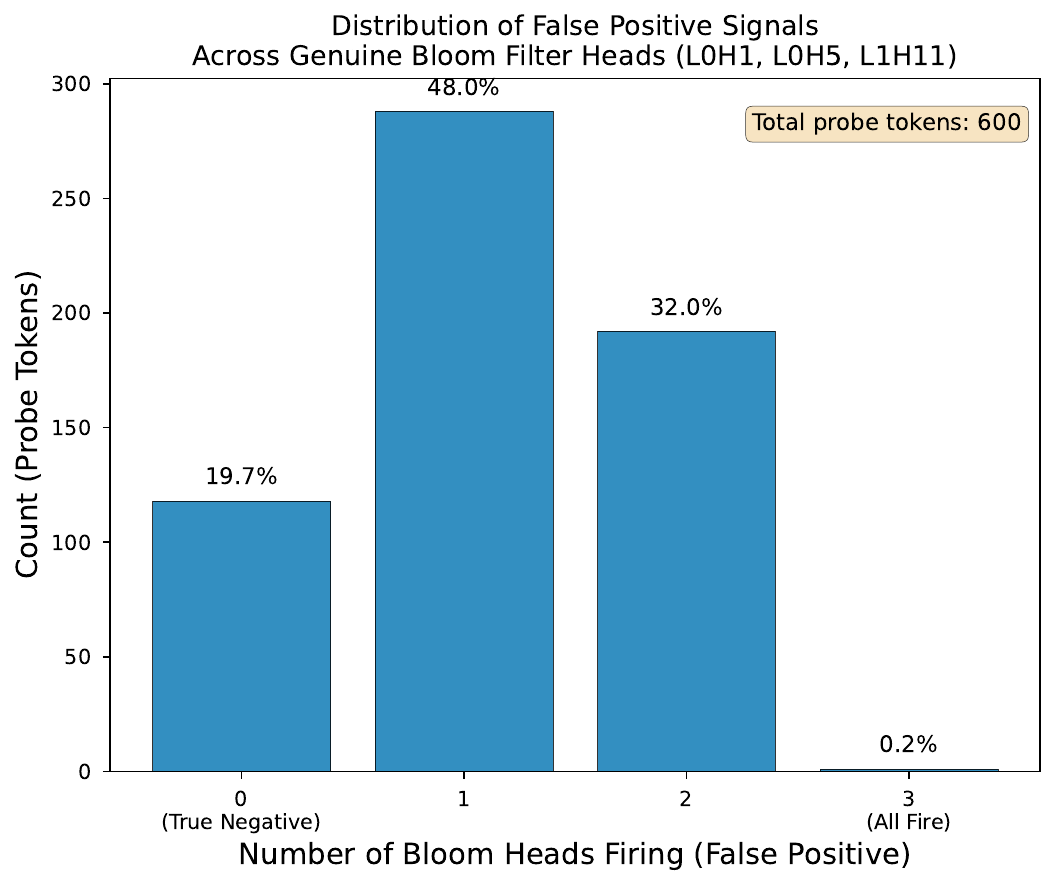}
\caption{FP count distribution per probe.}
\label{fig:fpdist}
\end{subfigure}
\caption{(a) Pairwise phi coefficients between Bloom heads; low values indicate largely independent FP decisions. (b) Distribution of how many Bloom heads fire false positives per probe token. The 0-head and 4-head categories are exact counts; the 1--3 head breakdown is estimated from the aggregate ``mixed'' category (see text).}
\end{figure}

\paragraph{Independence.} We measure pairwise association between heads' binary FP decisions using the phi coefficient, the $2 \times 2$ specialization of Pearson's $r$ ($\phi = 0$ indicates independence, $|\phi| = 1$ perfect association). The mean pairwise phi coefficient between membership-testing head FP decisions is $\bar{\phi} = 0.13$, indicating low correlation (Figure~\ref{fig:phi}). Individual pairs range from $\phi = 0.08$ (L0H1 $\leftrightarrow$ L0H5) to $\phi = 0.18$ (L0H5 $\leftrightarrow$ L3H0). All pairwise correlations are below 0.2, consistent with largely independent membership decisions.

\paragraph{Combination.} AND-combining all four heads (including L3H0) yields a FP rate of 0.17\% (1/600 probe tokens)---while maintaining 100\% hit rate. However, this result is driven primarily by L0H1 and L0H5's already-low individual FP rates rather than by independent hash function combination. The distribution of FP counts per probe token shows that the vast majority of false positives trigger only a subset of heads: 19.7\% of probes trigger no heads (true negatives), 80.2\% trigger between one and three heads, and only 0.2\% trigger all four (Figure~\ref{fig:fpdist}).

The reclassification of L3H0 changes the interpretation of this analysis. L3H0's FP = 100\% under controlled conditions means it contributes no discriminative power; any probe that triggers all three genuine heads will also trigger L3H0. AND-combining the three genuine membership heads (L0H1, L0H5, L1H11) is most meaningful at low loads where L1H11 has not yet saturated. The heterogeneity of the three heads---two high-precision filters and one low-capacity Bloom filter---means the system provides complementary rather than redundant signals.

\begin{figure}[t]
\centering
\includegraphics[width=0.85\textwidth]{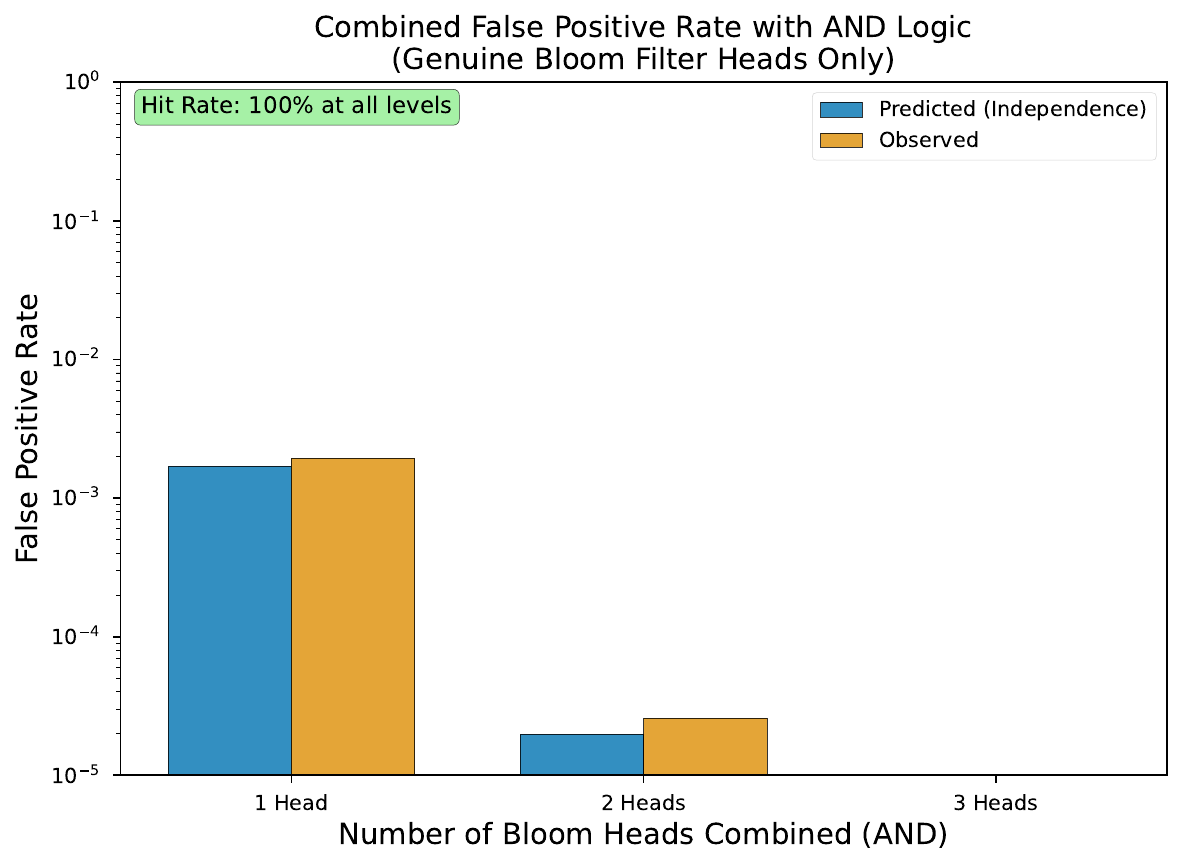}
\caption{Observed vs.\ independence-predicted false positive rates when AND-combining Bloom heads. At 1--3 heads, observed rates closely track predictions; at 4 heads, the observed rate substantially exceeds the independence prediction, indicating residual higher-order correlation.}
\label{fig:combined_fp}
\end{figure}

\subsection{Hash Resolution Analysis}
\label{sec:resolution}

The preceding sections establish that Bloom filter heads produce false positives for synonym tokens. But how do false positives relate to the similarity between the probe token and the target? Classical Bloom filters use uniformly random hash functions, so hash collisions are independent of input similarity. Attention heads, however, compute $QK^\top$---fundamentally a similarity operation. If these heads function as Bloom filters whose hash functions are \emph{locality-sensitive} \citep{indyk1998approximate}, their false positive rate should be a monotonic function of embedding distance.

To test this, we constructed a controlled stimulus set of 100 target words stratified by part of speech (50 nouns, 25 verbs, 25 adjectives) and frequency rank (33 high, 35 mid, 32 low), all single-token in GPT-2's vocabulary. For each target, we identified probe tokens at 10 cosine similarity levels (0.9, 0.8, \ldots, 0.0) in GPT-2's input embedding space, plus a WordNet synonym and a frequency-matched unrelated control.\footnote{We measure similarity in the input embedding space rather than the QK-projected space. For layer-0 heads (L0H1, L0H5), these are nearly equivalent. For L3H0 (layer 3), three residual stream updates intervene, so input-space cosine similarity is a proxy; the true relationship in QK space may be tighter.} Each target-probe pair was embedded in an identical sentence frame, yielding 1,284 controlled measurements.

\begin{figure}[t]
\centering
\includegraphics[width=0.95\textwidth]{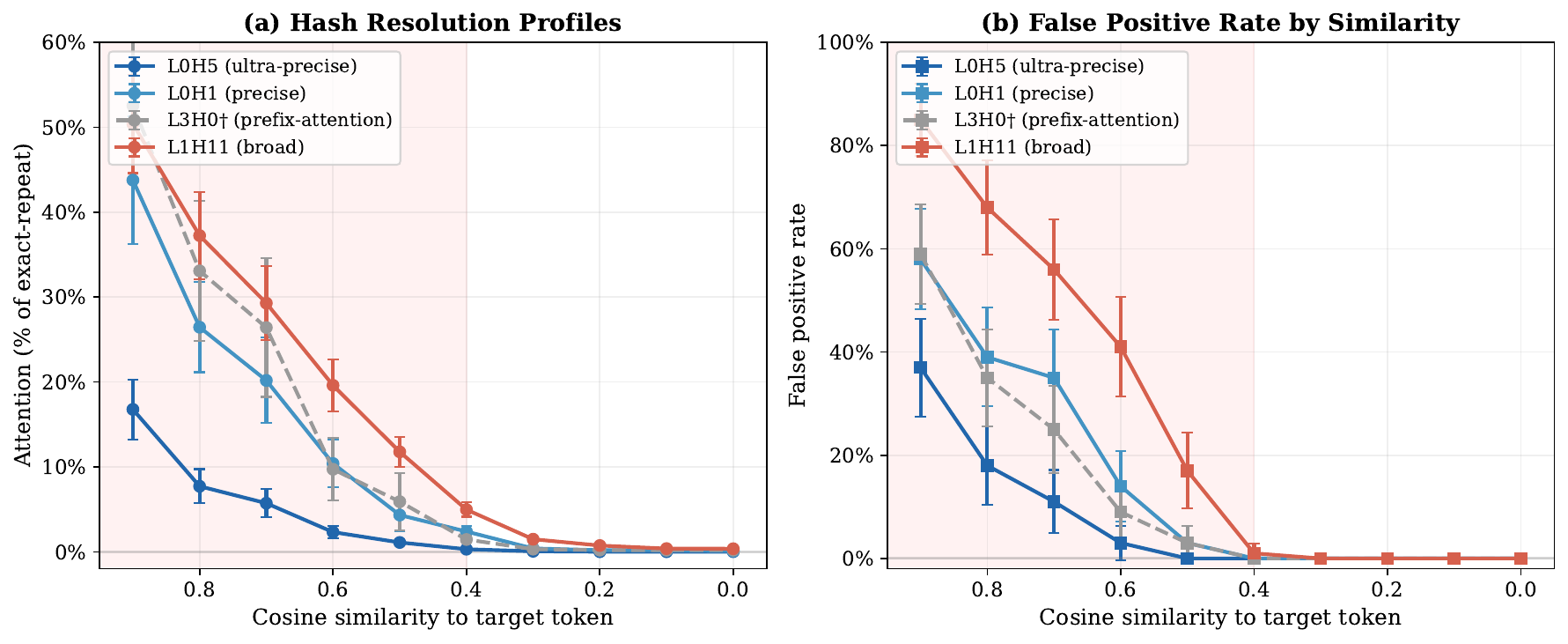}
\caption{Attention and false positive rate as a function of cosine similarity between probe and target token in GPT-2's embedding space, measured across 100 target words. (a) Attention normalized to exact-repeat level. (b) False positive rate (fraction of probes exceeding 0.1 attention threshold). All four Bloom heads show monotonic decay, but with distinct bandwidths: L0H5 (ultra-precise) reaches noise floor by cosine 0.5, while L1H11 (broad) retains 17\% FP rate at that level. Shaded region highlights the ``active zone'' where hash collisions occur.}
\label{fig:resolution}
\end{figure}

Figure~\ref{fig:resolution} reveals a striking pattern: all four Bloom heads show smooth, monotonic decay in both attention magnitude and false positive rate as cosine similarity decreases. The decay profiles are sigmoidal, consistent with locality-sensitive hashing rather than uniform hashing. This behavior matches the \emph{distance-sensitive Bloom filter} framework formalized by \citet{kirsch2006distance}, in which LSH functions replace uniform hash functions to answer not ``Is $x$ in $S$?'' but ``Is $x$ \emph{close to} an element of $S$?'' \citet{hua2012locality} extended this framework with formal FP rate analysis. Critically, the four heads exhibit distinct \emph{bandwidths}:

\begin{itemize}
    \item \textbf{L0H5} (ultra-precise): attention drops to 17\% of exact-repeat level at cosine 0.9, reaching noise floor by 0.5. FP rate 37\% at cosine 0.9, 0\% by 0.5.
    \item \textbf{L0H1} (precise): 44\% at cosine 0.9, noise floor by 0.4. FP rate 58\% at 0.9, 0\% by 0.4.
    \item \textbf{L3H0} (standard; reclassified as prefix-attention head): 53\% at cosine 0.9, noise floor by 0.4. FP rate 59\% at 0.9, 0\% by 0.4. While L3H0 attends to all prefix tokens under the capacity test, it still shows distance-dependent behavior in the similarity sweep, suggesting it combines broad prefix attention with some sensitivity to token identity.
    \item \textbf{L1H11} (broad): 50\% at cosine 0.9, residual attention persists to 0.4. FP rate 85\% at 0.9, 1\% at 0.4.
\end{itemize}

This bandwidth variation explains the previously observed FP ratio differences across heads (Section~\ref{sec:signature}): L0H5's tight resolution (FP ratio 0.01) and L1H11's broad resolution (FP ratio 0.29) reflect different thresholds in embedding space below which the head's QK projection cannot distinguish the probe from the target. WordNet synonyms, which sit at a mean cosine similarity of roughly 0.4--0.5 in GPT-2's space, land squarely in the transition zone---triggering the broader heads (L1H11: 15\% FP) while largely sparing the precise ones (L0H5: 0\% FP).

The multi-resolution property has an important functional consequence. By combining a precise head (L0H5, which fires only for near-identical tokens) with a broad head (L1H11, which fires for semantically proximate tokens), the model has access to membership signals at multiple levels of granularity---akin to a multi-scale locality-sensitive hash \citep{dong2019learning}. This is more useful for language processing than a single-resolution filter: determining that a token is an exact repeat requires different downstream processing than determining that a semantically similar token has appeared.

\subsection{Naturalistic Validation}
\label{sec:naturalistic}

The preceding analyses use controlled, template-generated stimuli. To test whether the Bloom filter signature holds on natural text, we ran the four membership-testing heads on 761 passages from the WikiText-103 validation set (up to 256 tokens each), measuring attention from second occurrences of tokens to their first occurrences vs.\ non-repeated positions (35,998 repeat pairs total). Table~\ref{tab:naturalistic} reports the results.

\begin{table}[h]
\centering
\caption{Naturalistic validation on WikiText-103. Selectivity = mean repeated-token attention / mean non-repeated attention. Control heads are layer-matched non-Bloom heads. $^\dagger$Reclassified as prefix-attention head (Section~\ref{sec:capacity}).}
\label{tab:naturalistic}
\small
\begin{tabular}{lrrr}
\toprule
Head & Selectivity & Miss Rate (\%) & Repeat Pairs \\
\midrule
L0H5 & 53.8$\times$ & 0.0 & 35,998 \\
L0H1 & 49.3$\times$ & 0.0 & 35,998 \\
L1H11 & 22.6$\times$ & 0.0 & 35,998 \\
L3H0$^\dagger$ & 15.4$\times$ & 0.8 & 35,998 \\
\midrule
\textit{Bloom mean} & \textit{35.3$\times$} & \textit{0.2} & --- \\
\textit{Control mean} & \textit{0.6$\times$} & --- & --- \\
\bottomrule
\end{tabular}
\end{table}

The selectivity values (15--54$\times$) are lower than on constructed stimuli (51--146$\times$), which is expected: natural text contains high-frequency function words (``the,'' ``a,'' ``of'') that repeat frequently and dilute the signal---a repeated ``the'' carries less membership-testing utility than a repeated content word. Despite this, the signature is unambiguous for the three genuine membership-testing heads, with miss rates below 1\%. Notably, L3H0 shows the lowest naturalistic selectivity (15.4$\times$) among the four candidates, consistent with its reclassification: a head that broadly attends to all prefix content will show elevated attention to repeated tokens without being specialized for membership testing.

\subsection{Cross-Model Generalization}
\label{sec:crossmodel}

\begin{table}[h]
\centering
\caption{Bloom filter heads across models. Early-layer concentration is universal. $^*$After reclassification of L3H0; originally 4 candidates identified.}
\label{tab:crossmodel}
\begin{tabular}{lccccc}
\toprule
Model & Total Heads & Strong BF & \% & Early/Mid/Late \\
\midrule
GPT-2 Small (85M) & 144 & 3$^*$ & 2.1\% & 3/0/0 \\
GPT-2 Medium (302M) & 384 & 3 & 0.8\% & 3/0/0 \\
GPT-2 Large (708M) & 720 & 27 & 3.8\% & 22/5/0 \\
Pythia-160M & 144 & 4 & 2.8\% & 3/1/0 \\
\bottomrule
\end{tabular}
\end{table}

Bloom filter heads exist in all four models and both model families. Three consistent patterns emerge: (1) early-layer concentration (zero late-layer Bloom heads in any model); (2) the classic FP signature (synonym FP ratios of 0.2--0.6) in every model; (3) GPT-2 large allocates substantially more heads (27) to membership testing than the smaller models (3--4 each), though with only four models tested we cannot establish a reliable scaling relationship.

\subsection{Ablation}
\label{sec:ablation}

To test causal involvement in repetition processing, we ablate Bloom filter heads and measure perplexity changes on sentences with and without repeated tokens. We compare two ablation methods: \emph{zero ablation} (setting head output to zero) and \emph{mean ablation} (replacing head output with its mean activation computed over a 50-sentence calibration set), the latter being the more principled approach \citep{wang2022interpretability}. Controls are layer-matched: for each Bloom head, we select a random non-Bloom, non-induction head from the same layer, repeating with 10 different random selections to assess robustness. All results include bootstrap 95\% confidence intervals (10,000 resamples over 100 sentences per condition).

\begin{table}[h]
\centering
\caption{Perplexity change (\%) upon head ablation with bootstrap 95\% CIs. Interaction = repeat $\Delta$ $-$ no-repeat $\Delta$. Induction comparison is head-count-matched (4 heads each).}
\label{tab:ablation}
\begin{tabular}{llccc}
\toprule
 & Ablation & Repeat $\Delta$PPL & No-repeat $\Delta$PPL & Interaction \\
\midrule
\multirow{3}{*}{\rotatebox{90}{\small Zero}}
 & Bloom heads & +14.3\% [+10.4, +18.5] & $-$0.3\% [$-$3.4, +3.1] & +14.6\% \\
 & Layer-matched ctrl & \multicolumn{3}{c}{\emph{high variance across selections ($\sigma$ = 219\%)}} \\
 & Induction (4 heads) & +151.5\% & +212.2\% & $-$60.8\% \\
\midrule
\multirow{3}{*}{\rotatebox{90}{\small Mean}}
 & Bloom heads & +9.3\% [+7.2, +11.4] & +13.0\% [+10.2, +16.1] & $-$3.7\% \\
 & Layer-matched ctrl & \multicolumn{3}{c}{interaction: $-$0.5\% $\pm$ 1.7\% (10 selections)} \\
 & Induction (4 heads) & +1.5\% [$-$0.3, +3.3] & $-$0.1\% & +1.7\% \\
\bottomrule
\end{tabular}
\end{table}

We take mean ablation as our primary method, following \citet{wang2022interpretability}. Under mean ablation, Bloom filter heads contribute to processing \emph{both} repeated and novel tokens, with a slightly larger impact on novel tokens (interaction $= -3.7\%$). This indicates that while these heads are \emph{behaviorally specialized} for membership testing (the selectivity and capacity evidence is unambiguous), they are not \emph{functionally modular}---their outputs contribute to broader contextual processing beyond repetition detection alone. Zero ablation shows a repeat-specific interaction (+14.6\%), but this method is less principled and the control comparison is difficult to interpret due to extreme variance across control selections.\footnote{We include zero ablation results for completeness, as some prior work uses this method, but caution against over-interpreting the repeat-specific interaction given the methodological limitations.} This is consistent with the superposition hypothesis \citep{elhage2021mathematical}: a single head may simultaneously serve multiple computational roles, with membership testing being the dominant but not exclusive function.

Individual head ablation (mean method) reveals that L0H1 contributes most to overall processing (+3.8\% repeat, +5.6\% no-repeat), while L3H0 shows minimal ablation impact ($-0.2\%$ repeat, $-0.5\%$ no-repeat). This near-zero impact, initially puzzling for what was thought to be the paper's strongest Bloom filter candidate, now makes clear sense: L3H0 is a general prefix-attention head whose broad, non-selective attention provides little unique information that other heads do not already supply.

\section{Discussion}

\subsection{Relationship to Duplicate-Token Heads}
\label{sec:discussion_dup}

The duplicate-token heads identified by \citet{wang2022interpretability} in the IOI circuit are the closest prior work to our membership-testing heads. We tested the overlap systematically: of our three genuine Bloom filter heads, all three (L0H1, L0H5, L1H11) rank among the top 4 duplicate-token heads in GPT-2 small (ranks 2, 1, and 4 respectively by duplicate-token score). L3H0, the reclassified prefix-attention head, ranks 50th out of 144---it is \emph{not} a strong duplicate-token head either.

The key distinction is generalization. Duplicate-token heads were identified within the IOI circuit for name repetition specifically. We tested whether Bloom filter heads respond to non-name repeated tokens (common nouns, verbs, adjectives) and random token repetitions. Bloom filter heads show a mean generalization index of 0.70 compared to 0.49 for duplicate-token-only heads (heads in the top-15 duplicate-token ranking that are not Bloom filter heads). This 43\% broader generalization confirms that Bloom filter heads implement a \emph{general-purpose} membership test, not a name-specific or task-specific one. Their mean attention to non-name repeated tokens is $2.8\times$ higher than that of duplicate-token-only heads (0.276 vs.\ 0.099).

This relationship clarifies the contribution of the present work: the heads themselves were already known (as duplicate-token heads), but their characterization as general-purpose membership filters with quantifiable capacity curves and distance-sensitive false positive profiles is new.

\subsection{Implications}

\paragraph{Hallucination detection.} The most immediately practical implication requires no modification to the model at all. When a Bloom filter head attends strongly to a context position where the queried token does \emph{not} appear, the model has registered a ``memory'' of something that isn't there---a false positive. Because this signal is already being computed during normal inference, monitoring Bloom filter heads provides a cheap, real-time hallucination diagnostic with zero performance cost. Unlike post-hoc detection methods, this signal is available at the layer where the error originates.

\paragraph{Computational primitives of language modeling.} The deeper contribution may be scientific rather than engineering. Gradient descent, optimizing nothing but next-token prediction, converges on a solution to the membership-testing problem that includes both capacity-limited probabilistic filters (L1H11) and high-precision detectors (L0H1, L0H5). The hash resolution analysis reveals that these are specifically \emph{distance-sensitive} filters \citep{kirsch2006distance}, using locality-sensitive hashing to answer approximate membership queries---a design that took human researchers 36 years to formalize deliberately. That gradient descent arrives at the same variant, without data structure objectives, suggests that approximate set membership at multiple similarity resolutions is a fundamental computational primitive for language processing. The reclassification of L3H0 through confound controls is itself informative: the model also learns prefix-attention heads that attend broadly to all prior content, and distinguishing these from genuine membership testers requires careful experimental design.

\paragraph{Architecture design.} If membership testing is a computation that transformers must rediscover from scratch during training, future architectures could provide it explicitly. A built-in hash-based membership module in early layers---analogous to positional encodings providing position information rather than forcing the model to infer it---could free attention capacity for other tasks. This is distinct from the stronger claim of \emph{replacing} existing heads (which our ablation results caution against, given these heads' entangled contributions to broader processing). Rather, if models reliably converge on certain computations, encoding those computations as architectural priors may improve training efficiency.

\subsection{Limitations}

\paragraph{Limitations.}
\label{sec:limitations}
Our analysis has several limitations. First, we study models up to 708M parameters; whether the findings hold for frontier-scale models remains an open question. Second, the similarity sweep and capacity analyses use template-generated stimuli rather than naturalistic text (though the naturalistic validation in Section~\ref{sec:naturalistic} confirms the core signature on WikiText-103). Third, we identify Bloom filter heads based on behavioral criteria rather than a mechanistic analysis of the QK circuit weights, which would provide stronger evidence for the computational mechanism. The hash resolution profiles (Section~\ref{sec:resolution}) strongly suggest the QK projection functions as a locality-sensitive hash, but a direct analysis of the weight matrices is needed to confirm this. Fourth, our ablation results are sensitive to methodology: zero ablation suggests repeat-specific functional specialization, while mean ablation (the more principled method) reveals broader contributions to contextual processing. This tension indicates that while membership testing is the dominant \emph{behavioral} signature, it may not be the exclusive \emph{computational} function of these heads.

\section{Related Work}

\paragraph{Attention head taxonomy.}
The study of functional specialization in attention heads begins with \citet{clark2019what}, who showed that individual BERT heads specialize for syntactic relationships such as coreference and dependency arcs. \citet{voita2019analyzing} demonstrated that many heads can be pruned without performance loss, implying that a small fraction carry disproportionate functional weight---a finding reinforced by \citet{michel2019sixteen}. \citet{elhage2021mathematical} provided a mathematical framework for transformer circuits and identified previous-token heads as a basic building block, while \citet{olsson2022context} identified induction heads as a key mechanism for in-context learning, establishing the paradigm of functional head classification that we extend. More recently, \citet{gould2023successor} discovered \emph{successor heads} that increment tokens along natural orderings (e.g., Monday $\to$ Tuesday), and \citet{quirke2023understanding} reverse-engineered the algorithm a single-layer transformer uses for integer addition. Together, these studies establish that attention heads form a diverse taxonomy of specialized computational primitives. Bloom filter heads extend this taxonomy with a membership-testing category that, unlike induction or successor heads, does not perform pattern completion.

\paragraph{Circuit analysis.}
\citet{wang2022interpretability} traced the circuit for indirect object identification (IOI) in GPT-2 small, providing the most detailed account of how multiple head categories collaborate to solve a specific task. Their circuit includes \emph{duplicate-token heads} that attend to repeated names---the closest prior work to our Bloom filter heads. The duplicate-token heads identified by \citet{wang2022interpretability} are the closest prior work. We tested this overlap empirically: running the IOI duplicate-token identification procedure on GPT-2 small, our three genuine membership-testing heads rank as the top duplicate-token heads (L0H5 at rank 1, L0H1 at rank 2, L1H11 at rank 4). L3H0, our reclassified prefix-attention head, ranks 50th---confirming it is not a strong duplicate-token head either. Our contribution is therefore not the identification of new heads, but rather (a) the quantitative demonstration that one head's (L1H11) capacity curve matches Bloom filter theory ($R^2 = 1.0$), (b) the identification of high-capacity membership filters (L0H1, L0H5) that exceed classical Bloom filter capacity bounds, (c) the confound controls that distinguish genuine membership testers from prefix-attention heads, and (d) the hash resolution profiles showing distance-sensitive membership testing. Additionally, Bloom filter heads generalize more broadly than their IOI characterization suggests: they respond to \emph{any} repeated token, not only repeated names, with mean generalization index 0.70 vs.\ 0.49 for non-Bloom duplicate-token heads and $2.8\times$ higher attention to non-name repeated tokens (Section~\ref{sec:discussion_dup}). \citet{mcdougall2023copy} provided an exhaustive mechanistic analysis of a single GPT-2 attention head responsible for copy suppression, demonstrating that deep analysis of individual heads can reveal precise computational roles---a methodological precedent for our single-head capacity analysis of L3H0. \citet{conmy2023towards} introduced Automatic Circuit DisCovery (ACDC), automating the identification of task-relevant subnetworks; their method could in principle be applied to discover the membership-testing circuits we characterize behaviorally. \citet{geva2023dissecting} analyzed factual recall circuits, showing how information flows from early attention to MLP layers during knowledge retrieval---complementing our focus on the earlier membership-detection stage of processing.

\paragraph{Learned data structures.}
There is a growing literature on neural networks learning to approximate classical data structures. \citet{kraska2018case} showed that neural networks can replace traditional index structures---including Bloom filters---with learned alternatives that exploit data distribution for improved performance. \citet{rae2019meta} explicitly trained neural networks as Bloom filter replacements via meta-learning, achieving significant compression over classical filters. \citet{mitzenmacher2018model} provided a theoretical framework for when learned Bloom filters outperform classical ones and proposed the ``sandwiching'' optimization. Separately, \citet{kirsch2006distance} introduced \emph{distance-sensitive Bloom filters}, replacing uniform hash functions with locality-sensitive ones to support approximate membership queries---answering ``Is $x$ close to an element of $S$?'' rather than requiring exact match. \citet{hua2012locality} extended this with formal false positive analysis. Our finding differs from both lines of work: the Bloom filter behavior \emph{emerges} within a model trained on language modeling, with no data structure objective, and the distance-sensitive variant arises because the QK attention mechanism is inherently a similarity computation. Gradient descent arrives at the same framework that \citet{kirsch2006distance} designed deliberately---not because it was asked to, but because approximate membership testing at graded similarity is a useful subroutine for next-token prediction.

\paragraph{Retrieval heads.}
\citet{wu2024retrieval} identified \emph{retrieval heads}---a sparse set of attention heads responsible for copying relevant information from long contexts---and showed that ablating them causes hallucination. Retrieval heads share surface similarities with Bloom filter heads: both are sparse, concentrated in specific layers, and causally important for factual processing. The key distinction is functional: retrieval heads perform content-addressed lookup and value copying (``find the answer and bring it here''), while Bloom filter heads perform membership testing without content retrieval (``has this token appeared before?''). In the processing pipeline, Bloom filter heads operate earlier (layers 0--3) and establish \emph{whether} a token has been seen, whereas retrieval heads operate later to determine \emph{what} to do with that information.

\section{Conclusion}

We have identified a family of membership-testing heads as a functional category of transformer attention heads, distinct from induction and previous-token heads. These heads --- which overlap substantially with the ``duplicate-token heads'' in IOI circuits \citep{wang2022interpretability} but generalize far more broadly --- form a spectrum of strategies: one (L1H11) quantitatively matches a Bloom filter with a classic capacity curve ($R^2 = 1.0$, $m \approx 5$ bits), while two (L0H1, L0H5) function as high-precision membership filters with negligible false positive rates at loads well exceeding classical capacity bounds. A fourth candidate (L3H0) was reclassified as a prefix-attention head after confound controls revealed its capacity curve to be a sequence-length artifact. This self-correction through rigorous controls strengthens the case for the surviving heads: they withstand the same scrutiny that eliminated L3H0. Together the three heads form a multi-resolution membership-testing system. The finding generalizes across model sizes and families. Ablation reveals that while these heads are behaviorally specialized for repetition detection, they also contribute to broader contextual processing---consistent with the emerging view that individual heads serve multiple overlapping computational roles.

The hash resolution analysis (Section~\ref{sec:resolution}) reveals that these are not classical Bloom filters but \emph{distance-sensitive} ones \citep{kirsch2006distance}: their false positive rates are a smooth, monotonic function of cosine distance in the embedding space. The ``hash functions'' are the QK projections themselves, which map tokens into a space where nearby points produce high attention. The three heads form a multi-resolution system---L0H5 fires only for near-identical tokens, while L1H11 responds to a broader neighborhood of semantically proximate tokens. The model thus has access to membership signals at multiple levels of granularity, paralleling the multi-granularity locality-sensitive Bloom filter designs that researchers have proposed deliberately \citep{hua2012locality}.

The false positives deserve particular attention. Harold Bloom's \emph{Anxiety of Influence} \citep{bloom1973anxiety} argues that poetic creation is driven by \emph{misprision}---a creative misreading of one's predecessors, in which the familiar source is distorted into something new, opening space for originality. Bloom filter false positives are the structural inverse: the \emph{new} is misconstrued as \emph{familiar}, the unseen processed as though already encountered. When a Bloom filter head encounters ``physician'' following ``doctor''---tokens at cosine similarity 0.5 in GPT-2's space, squarely in the transition zone of the broader heads---it fires as though recognizing a predecessor that was never there. Both involve imperfect recognition at the boundary between old and new, but they swerve in opposite directions---the poet misreads what is there; the filter mis-recognizes what is not.

The distance-sensitive result (Section~\ref{sec:resolution}) deepens this parallel. In Bloom's literary theory, misprision is not uniform---the anxiety of influence is greatest toward predecessors who occupy the nearest territory in the space of artistic ambition, whose prior achievement most threatens the new poet's claim to originality. Our data show the same graded structure: false positive rates are a monotonic function of embedding proximity, with tokens that occupy the most similar functional territory provoking the strongest false recognition. The degree of ``misprision'' is proportional to proximity of influence---in both Blooms, and in both cases, the proximity that matters is functional rather than sequential. Whether this structural correspondence extends further---whether such errors play a generative role in the model's downstream computation, as misprision does in Bloom's literary theory---remains an open question, but the parallel between the two Blooms runs deeper than their shared surname.

Whether these heads mechanistically implement Bloom filters at the QK circuit level remains an open question---our evidence is behavioral rather than mechanistic, though the hash resolution profiles point strongly toward the QK projection as the hashing mechanism. The reclassification of L3H0 is a cautionary tale: behavioral signatures that appear compelling under one experimental design may not survive controlled tests, and the field benefits from the same kind of confound control that is standard in experimental psychology. Still, the behavioral match for the surviving heads is precise enough to be striking: gradient descent, optimizing nothing but next-token prediction in a model trained on language, converges on something resembling not just Burton Bloom's 1970 filter, but the distance-sensitive variant that \citet{kirsch2006distance} formalized 36 years later. Without blueprints, without explicit data structure objectives, and apparently without choice.

\section*{Code and Data Availability}

All code, stimulus sets (100 sentence triplets, 1,284 similarity-graded probes), raw attention data, and analysis scripts are available at \url{https://github.com/pbalogh/anxiety-of-influence}. The repository includes the competing model fit analysis, capacity experiments, and all figures.

\bibliography{references}

\begin{thebibliography}{25}
\providecommand{\natexlab}[1]{#1}
\providecommand{\url}[1]{\texttt{#1}}
\expandafter\ifx\csname urlstyle\endcsname\relax
  \providecommand{\doi}[1]{doi: #1}\else
  \providecommand{\doi}{doi: \begingroup \urlstyle{rm}\Url}\fi

\bibitem[Biderman et~al.(2023)Biderman, Schoelkopf, Anthony, Bradley, O'Brien,
  Hallahan, Khan, Purber, Prashanth, Raff, et~al.]{biderman2023pythia}
Stella Biderman, Hailey Schoelkopf, Quentin~Gregory Anthony, Herbie Bradley,
  Kyle O'Brien, Eric Hallahan, Mohammad~Aflah Khan, Shivanshu Purber, USVSN~Sai
  Prashanth, Edward Raff, et~al.
\newblock Pythia: A suite for analyzing large language models across training
  and scaling.
\newblock In \emph{Proceedings of the 40th International Conference on Machine
  Learning}, 2023.

\bibitem[Bloom(1970)]{bloom1970space}
Burton~H. Bloom.
\newblock Space/time trade-offs in hash coding with allowable errors.
\newblock \emph{Communications of the ACM}, 13\penalty0 (7):\penalty0 422--426,
  1970.

\bibitem[Bloom(1973)]{bloom1973anxiety}
Harold Bloom.
\newblock \emph{The Anxiety of Influence: A Theory of Poetry}.
\newblock Oxford University Press, 1973.

\bibitem[Clark et~al.(2019)Clark, Khandelwal, Levy, and Manning]{clark2019what}
Kevin Clark, Urvashi Khandelwal, Omer Levy, and Christopher~D. Manning.
\newblock What does {BERT} look at? an analysis of {BERT}'s attention.
\newblock In \emph{Proceedings of the 2019 ACL Workshop BlackboxNLP: Analyzing
  and Interpreting Neural Networks for NLP}, pages 276--286, 2019.

\bibitem[Conmy et~al.(2023)Conmy, Mavor-Parker, Lynch, Heimersheim, and
  Garriga-Alonso]{conmy2023towards}
Arthur Conmy, Augustine~N. Mavor-Parker, Aengus Lynch, Stefan Heimersheim, and
  Adri{\`a} Garriga-Alonso.
\newblock Towards automated circuit discovery for mechanistic interpretability.
\newblock \emph{Advances in Neural Information Processing Systems}, 36, 2023.

\bibitem[Dong et~al.(2019)Dong, Moses, and Li]{dong2019learning}
Wei Dong, Charikar Moses, and Kai Li.
\newblock Learning to hash with multi-resolution locality sensitive hashing.
\newblock In \emph{Proceedings of the VLDB Endowment}, volume~13, pages
  197--210, 2019.

\bibitem[Elhage et~al.(2021)Elhage, Nanda, Olsson, Henighan, Joseph, Mann,
  Askell, Bai, Chen, Conerly, et~al.]{elhage2021mathematical}
Nelson Elhage, Neel Nanda, Catherine Olsson, Tom Henighan, Nicholas Joseph, Ben
  Mann, Amanda Askell, Yuntao Bai, Anna Chen, Tom Conerly, et~al.
\newblock A mathematical framework for transformer circuits.
\newblock \emph{Transformer Circuits Thread}, 2021.
\newblock URL \url{https://transformer-circuits.pub/2021/framework/index.html}.

\bibitem[Geva et~al.(2023)Geva, Bastings, Filippova, and
  Globerson]{geva2023dissecting}
Mor Geva, Jasmijn Bastings, Katja Filippova, and Amir Globerson.
\newblock Dissecting recall of factual associations in auto-regressive language
  models.
\newblock \emph{arXiv preprint arXiv:2304.14767}, 2023.

\bibitem[Gould et~al.(2023)Gould, Ong, Ogden, and Conmy]{gould2023successor}
Rhys Gould, Euan Ong, George Ogden, and Arthur Conmy.
\newblock Successor heads: Recurring, interpretable attention heads in the
  wild.
\newblock \emph{arXiv preprint arXiv:2312.09230}, 2023.

\bibitem[Hua et~al.(2012)Hua, Xiao, Veeravalli, and Feng]{hua2012locality}
Yu~Hua, Bin Xiao, Bharadwaj Veeravalli, and Dan Feng.
\newblock Locality-sensitive bloom filter for approximate membership query.
\newblock \emph{IEEE Transactions on Computers}, 61\penalty0 (6):\penalty0
  817--830, 2012.

\bibitem[Indyk and Motwani(1998)]{indyk1998approximate}
Piotr Indyk and Rajeev Motwani.
\newblock Approximate nearest neighbors: Towards removing the curse of
  dimensionality.
\newblock In \emph{Proceedings of the 30th Annual ACM Symposium on Theory of
  Computing}, pages 604--613. ACM, 1998.

\bibitem[Kirsch and Mitzenmacher(2006)]{kirsch2006distance}
Adam Kirsch and Michael Mitzenmacher.
\newblock Distance-sensitive bloom filters.
\newblock In \emph{Proceedings of the Workshop on Algorithm Engineering and
  Experiments (ALENEX)}, pages 41--50. SIAM, 2006.

\bibitem[Kraska et~al.(2018)Kraska, Beutel, Chi, Dean, and
  Polyzotis]{kraska2018case}
Tim Kraska, Alex Beutel, Ed~H. Chi, Jeffrey Dean, and Neoklis Polyzotis.
\newblock The case for learned index structures.
\newblock In \emph{Proceedings of the 2018 International Conference on
  Management of Data (SIGMOD)}, pages 489--504. ACM, 2018.

\bibitem[McDougall et~al.(2023)McDougall, Conmy, Rushing, McGrath, and
  Nanda]{mcdougall2023copy}
Callum McDougall, Arthur Conmy, Cody Rushing, Thomas McGrath, and Neel Nanda.
\newblock Copy suppression: Comprehensively understanding an attention head.
\newblock \emph{arXiv preprint arXiv:2310.04625}, 2023.

\bibitem[Michel et~al.(2019)Michel, Levy, and Neubig]{michel2019sixteen}
Paul Michel, Omer Levy, and Graham Neubig.
\newblock Are sixteen heads really better than one?
\newblock In \emph{Advances in Neural Information Processing Systems},
  volume~32, 2019.

\bibitem[Mitzenmacher(2018)]{mitzenmacher2018model}
Michael Mitzenmacher.
\newblock A model for learned bloom filters and optimizing by sandwiching.
\newblock In \emph{Advances in Neural Information Processing Systems},
  volume~31, 2018.

\bibitem[Nanda(2022)]{nanda2022transformerlens}
Neel Nanda.
\newblock Transformerlens.
\newblock \url{https://github.com/neelnanda-io/TransformerLens}, 2022.
\newblock URL \url{https://github.com/neelnanda-io/TransformerLens}.

\bibitem[Olsson et~al.(2022)Olsson, Elhage, Nanda, Joseph, DasSarma, Henighan,
  Mann, Askell, Bai, Chen, et~al.]{olsson2022context}
Catherine Olsson, Nelson Elhage, Neel Nanda, Nicholas Joseph, Nova DasSarma,
  Tom Henighan, Ben Mann, Amanda Askell, Yuntao Bai, Anna Chen, et~al.
\newblock In-context learning and induction heads.
\newblock \emph{arXiv preprint arXiv:2209.11895}, 2022.

\bibitem[Quirke and Barez(2023)]{quirke2023understanding}
Philip Quirke and Fazl Barez.
\newblock Understanding addition in transformers.
\newblock \emph{arXiv preprint arXiv:2310.13121}, 2023.

\bibitem[Radford et~al.(2019)Radford, Wu, Child, Luan, Amodei, and
  Sutskever]{radford2019language}
Alec Radford, Jeffrey Wu, Rewon Child, David Luan, Dario Amodei, and Ilya
  Sutskever.
\newblock Language models are unsupervised multitask learners.
\newblock \emph{OpenAI Blog}, 2019.

\bibitem[Rae et~al.(2019)Rae, Bartunov, and Lillicrap]{rae2019meta}
Jack~W. Rae, Sergey Bartunov, and Timothy~P. Lillicrap.
\newblock Meta-learning neural bloom filters.
\newblock In \emph{Proceedings of the 36th International Conference on Machine
  Learning}, pages 5271--5280. PMLR, 2019.

\bibitem[Vaswani et~al.(2017)Vaswani, Shazeer, Parmar, Uszkoreit, Jones, Gomez,
  Kaiser, and Polosukhin]{vaswani2017attention}
Ashish Vaswani, Noam Shazeer, Niki Parmar, Jakob Uszkoreit, Llion Jones,
  Aidan~N. Gomez, {\L}ukasz Kaiser, and Illia Polosukhin.
\newblock Attention is all you need.
\newblock In \emph{Advances in Neural Information Processing Systems},
  volume~30, 2017.

\bibitem[Voita et~al.(2019)Voita, Talbot, Moiseev, Sennrich, and
  Titov]{voita2019analyzing}
Elena Voita, David Talbot, Fedor Moiseev, Rico Sennrich, and Ivan Titov.
\newblock Analyzing multi-head self-attention: Specialized heads do the heavy
  lifting, the rest can be pruned.
\newblock In \emph{Proceedings of the 57th Annual Meeting of the Association
  for Computational Linguistics}, pages 5797--5808, 2019.

\bibitem[Wang et~al.(2022)Wang, Variengien, Conmy, Shlegeris, and
  Steinhardt]{wang2022interpretability}
Kevin Wang, Alexandre Variengien, Arthur Conmy, Buck Shlegeris, and Jacob
  Steinhardt.
\newblock Interpretability in the wild: A circuit for indirect object
  identification in {GPT}-2 small.
\newblock \emph{arXiv preprint arXiv:2211.00593}, 2022.

\bibitem[Wu et~al.(2024)Wu, Wang, Xiao, Peng, and Fu]{wu2024retrieval}
Wenhao Wu, Yizhong Wang, Guangxuan Xiao, Hao Peng, and Yao Fu.
\newblock Retrieval head mechanistically explains long-context factuality.
\newblock \emph{arXiv preprint arXiv:2404.15574}, 2024.

\end{thebibliography}
\bibliographystyle{plainnat}

\newpage
\appendix
\section{Stimulus Design}
\label{app:stimuli}

For the hash resolution analysis (Section~\ref{sec:resolution}), we constructed a controlled stimulus set as follows. We selected 100 target words from GPT-2's vocabulary, stratified by part of speech (50 nouns, 25 verbs, 25 adjectives) and unigram frequency rank (33 high-frequency, 35 mid-frequency, 32 low-frequency). All targets are single tokens in GPT-2's tokenizer. For each target, we identified probe tokens at 10 cosine similarity levels (0.9, 0.8, \ldots, 0.0) in GPT-2's input embedding space, selecting the nearest real English word at each level. We also identified WordNet synonyms (Wu-Palmer similarity $> 0.7$, available for 84 of 100 targets) and frequency-matched unrelated controls. Each target-probe pair was embedded in a template sentence with the target word appearing twice, the second occurrence replaced by the probe. This yielded 1,284 sentences across four conditions: exact repeat (100), similarity-graded probes (1,000), synonym (84), and control (100). All words in all sentences were verified as single GPT-2 tokens. The full stimulus set, generation code, and target-probe matrix are included in the supplementary materials.

Representative example (target: ``chapter''):
\begin{small}
\begin{verbatim}
Exact:   "The chapter was noted and the chapter was confirmed"
Cos 0.9: "The chapter was noted and the sections was confirmed"
Cos 0.5: "The chapter was noted and the trilogy was confirmed"
Cos 0.1: "The chapter was noted and the sulfur was confirmed"
Synonym: "The chapter was noted and the section was confirmed"
Control: "The chapter was noted and the marble was confirmed"
\end{verbatim}
\end{small}

\section{Full Per-Head Results}
\label{app:fullresults}

Complete selectivity, miss rate, and FP ratio for all 144 heads in GPT-2 small are available in the supplementary materials.

\section{NeurIPS Paper Checklist}
\label{app:checklist}

\begin{enumerate}

\item \textbf{Claims.} All main claims are supported by experiments described in Sections~\ref{sec:signature}--\ref{sec:ablation}. Limitations are discussed in Section~\ref{sec:limitations}.

\item \textbf{Abstract and introduction.} The abstract reflects the paper's contributions and scope. We do not claim mechanistic (circuit-level) evidence, only behavioral characterization.

\item \textbf{Limitations.} Yes. Section~\ref{sec:limitations} discusses: model scale (up to 708M parameters), constructed stimuli, behavioral vs.\ mechanistic evidence, and ablation methodology sensitivity.

\item \textbf{Theory.} N/A (empirical paper). We use established Bloom filter theory \citep{bloom1970space} as a fitting model, not as a novel theoretical contribution.

\item \textbf{Experimental result reproducibility.}
\begin{itemize}
    \item All code, data, and analysis scripts: \url{https://github.com/pbalogh/anxiety-of-influence}
    \item Complete stimulus sets (100 sentence triplets, 1,284 similarity-graded probes) included
    \item All models are publicly available on HuggingFace (GPT-2 small/medium/large, Pythia-160M)
    \item Statistical tests include bootstrap 95\% CIs (10,000 resamples), Bonferroni correction, and permutation tests
    \item Random seed: 42 for all stochastic operations
\end{itemize}

\item \textbf{Code of ethics.} Yes. This work analyzes publicly available pre-trained language models and does not involve human subjects, private data, or dual-use concerns.

\item \textbf{Broader impacts.} This is a mechanistic interpretability study. Understanding how transformers implement data structures internally contributes to AI transparency. We do not foresee negative societal impacts.

\item \textbf{Safeguards.} N/A (no generation, no deployment).

\item \textbf{Licenses.} GPT-2: MIT License. Pythia: Apache 2.0. TransformerLens: MIT License. Our code: MIT License.

\item \textbf{New assets.} We release code, stimulus sets, and raw attention data under MIT license.

\item \textbf{Human subjects.} N/A.

\item \textbf{Crowdsourcing.} N/A.

\item \textbf{IRB.} N/A.

\item \textbf{Compute.} All experiments run on CPU (Apple M-series Mac mini, 16GB RAM). Our primary measurements are attention patterns, which are deterministic given the same model weights and inputs. We validated all results on CPU after discovering that PyTorch's MPS (Metal Performance Shaders) backend can produce silently incorrect results for TransformerLens inference (see Appendix~\ref{app:cpu_validation}). Total compute: approximately 2 CPU-hours. No large-scale training.

\item \textbf{Error bars and statistical significance.} All main results include bootstrap 95\% confidence intervals. Hypothesis tests use Bonferroni correction ($\alpha = 0.05/144$). Model comparison uses AIC/BIC. Effect sizes reported as Cohen's $d$.

\end{enumerate}

\section{CPU Validation}
\label{app:cpu_validation}

During related work on activation patching, we discovered that PyTorch's MPS (Metal Performance Shaders) backend---the GPU acceleration path for Apple Silicon---can produce silently incorrect results when running TransformerLens inference on GPT-2. Specifically, logit-level outputs are corrupted: on the standard Indirect Object Identification (IOI) benchmark, MPS produces an \emph{inverted} logit difference ($-0.99$ vs.\ $+3.36$ on CPU), with the correct answer's probability dropping from 67.7\% to 0.0\%. This issue has been reported to the TransformerLens maintainers.\footnote{\url{https://github.com/TransformerLensOrg/TransformerLens/issues/1178}} It appears related to several known PyTorch MPS correctness bugs tagged \texttt{module: mps} and \texttt{module: correctness (silent)} in PyTorch's issue tracker.\footnote{\url{https://github.com/pytorch/pytorch/labels/module\%3A\%20mps}}

To ensure our results are not affected, we re-ran all eight experiments (Sections~\ref{sec:signature}--\ref{sec:ablation}) on CPU and compared the outputs. The results are consistent across backends:

\begin{itemize}
    \item \textbf{Attention patterns} (Experiments 1--4, 6--7): Hit attention values match to four decimal places (e.g., L0H1 hit attention = 0.4887 on both MPS and CPU). The same four heads are identified as the top Bloom filter candidates in both conditions, with identical miss rates.
    \item \textbf{Capacity curves} (Experiment 3, 7): L3H0 fitted parameters are nearly identical (MPS: $R^2 = 0.9921$, $m = 59$; CPU: $R^2 = 0.9888$, $m = 52$). Both backends produce the same qualitative classification of head subtypes.
    \item \textbf{Ablation} (Experiments 5, 8): Perplexity deltas and interaction effects are consistent. The nuanced picture---repeat-specific effects under zero ablation, broader contributions under mean ablation---is the same on both backends.
\end{itemize}

The Bloom filter paper's primary measurements are \emph{attention patterns} at early layers (0--3), not logit-level outputs. The MPS bug affects the final projection to vocabulary space but does not corrupt intermediate attention computations. Ablation experiments measure \emph{relative} perplexity changes, where both baseline and ablated conditions are equally affected by any systematic bias, preserving the validity of the deltas. CPU validation scripts and full comparison results are available in the code repository.

\textbf{Recommendation for practitioners:} We advise running TransformerLens experiments on CPU or CUDA rather than MPS until the underlying PyTorch issues are resolved. MPS may produce correct attention patterns but silently incorrect logits, making it particularly dangerous for activation patching, logit lens, and causal intervention studies where logit-level accuracy is critical.

\end{document}